\documentclass[letterpaper, 10 pt, conference]{ieeeconf}  % Comment this line out if you need a4paper
\IEEEoverridecommandlockouts                              % This command is only needed if 
\usepackage{comment}

\usepackage{graphics} % for pdf, bitmapped graphics files
\usepackage{epsfig} % for postscript graphics files
\usepackage{mathptmx} % assumes new font selection scheme installed
\usepackage{times} % assumes new font selection scheme installed
\usepackage{amsmath} % assumes amsmath package installed
\usepackage{amssymb}  % assumes amsmath package installed
\usepackage{soul,color}
\usepackage{todonotes}
\usepackage{textcomp}
\title{\LARGE \bf
Automated robotic intraoperative ultrasound for brain surgery}

\author{Michael Dyck*$^{1,2}$, Alistair Weld*$^{3}$, Julian Klodmann$^{1}$, Alexander Kirst$^{1}$, Giulio Anichini$^{4}$, \\Luke Dixon$^{4}$, Sophie Camp$^{4}$, Stamatia Giannarou$^{3}$, Alin Albu-Schäffer$^{1,2}$% <-this % stops a space
\thanks{$^{1}$Institute of Robotics and Mechatronics, German Aerospace Center, $^{2}$Department of Informatics, Technical University of Munich, $^{3}$Hamlyn Centre for Robotic Surgery, Imperial College London, $^{4}$Department of Neurosurgery, Charing Cross Hospital, Imperial College London, UK}%
\thanks{*These authors contributed equally to the work - michael.dyck@dlr.de,  a.weld20@imperial.ac.uk. This project was supported by the International Graduate School of Science and Engineering (IGSSE); TUM Graduate School, and the UK Research and Innovation (UKRI) Centre for Doctoral Training in AI for Healthcare (EP/S023283/1), the Royal Society (URF$\setminus$R$\setminus$2 01014]), the NIHR Imperial Biomedical Research Centre. The authors want to thank \textit{ImFusion GmbH}, and especially Dr. Marco Esposito, for providing us an academic license to their software and constant support throughout our research.}
}

\begin{document}

\maketitle
\thispagestyle{empty}
\pagestyle{empty}

%%%%%%%%%%%%%%%%%%%%%%%%%%%%%%%%%%%%%%%%%%%%%%%%%%%%%%%%%%%%%%%%%%%%%%%%%%%%%%%%
%\begin{abstract}

% This electronic document is a ÒliveÓ template. The various components of your paper [title, text, heads, etc.] are already defined on the style sheet, as illustrated by the portions given in this document.

% \end{abstract}

%%%%%%%%%%%%%%%%%%%%%%%%%%%%%%%%%%%%%%%%%%%%%%%%%%%%%%%%%%%%%%%%%%%%%%%%%%%%%%%%
\section{Introduction}
%\vspace{-0.1cm}
During brain tumour resection, localising cancerous tissue and delineating healthy and pathological borders is challenging, even for experienced neurosurgeons and neuroradiologists \cite{Dixon2022IntraoperativeUI}. Intraoperative imaging is commonly employed for determining and updating surgical plans in the operating room. Ultrasound (US) has presented itself a suitable tool for this task, owing to its ease of integration into the operating room and surgical procedure. However, widespread establishment of this tool has been limited because of the difficulty of anatomy localisation and data interpretation.

Publications on various applications of robotic US scanning have been produced in recent years. Robotic ultrasound scanning of the spine is a well-posed tracking task. %\cite{uligoj2021RobUStAnAR} uses manual placement of the probe on the base of the spine, and with ultrasound feedback, tracks the spinous process.
\cite{Victorova2021FollowTC} uses a semantic segmentation network to segment the spine, and in conjunction with a depth map, generates a robotic trajectory. \cite{Ma2021AutonomousST} presents an automated method for lung diagnosis, using a pose estimation network to identify regions on the chest for path defining. 
\cite{jiang} uses image quality optimization and interaction force adjustment to optimize probe pose and surface coupling. %Usually, purely autonomous execution is considered, and external imaging and tracking systems are used for surface reconstruction and imaging registration.

In this work, we present a robotic framework designed and tested on a soft-tissue-mimicking brain phantom, simulating intraoperative US (iUS) scanning during brain tumour surgery. Our work builds on what was introduced in \cite{Weld2022CollaborativeRU,dyck2022collab}. 

%We focus on the integration of software and hardware components and the development of an autonomous object localisation and surface reconstruction of a complex medical phantom. This enables the seamless application of the robotic US controller introduced in \cite{dyck2022}. The resulting versatility of our robotic platform makes it applicable to a wide range of US applications, being especially suitable for scanning of soft tissue with complex geometries.

\begin{figure}[t]
  \centering
  \fontsize{9}{9}
  \selectfont
  \includegraphics[width=0.8\linewidth]{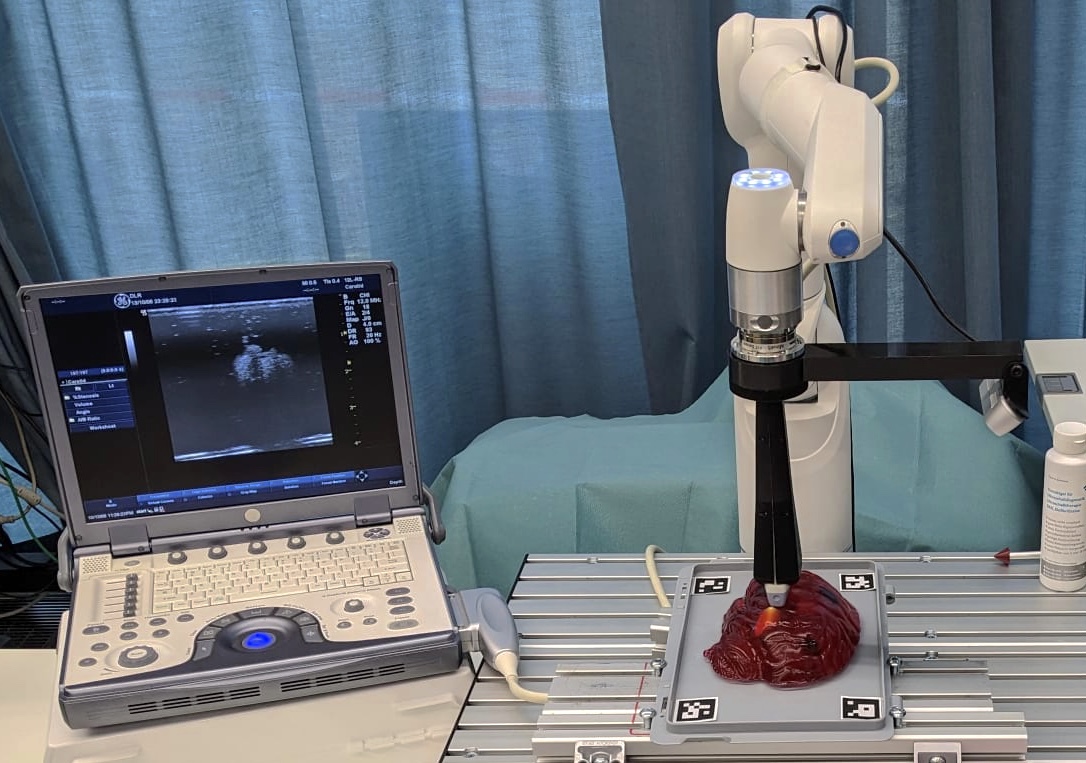}
  \caption{Experimental setup showing the robotic arm with it's attached probe and camera holder. Scanning a custom-made brain phantom.}
  \label{fig:setup}
\end{figure}

\section{Methods}
\begin{figure*}[!ht]
  \centering
  \fontsize{9}{9}
  \selectfont
  \includegraphics[width=1.\textwidth]{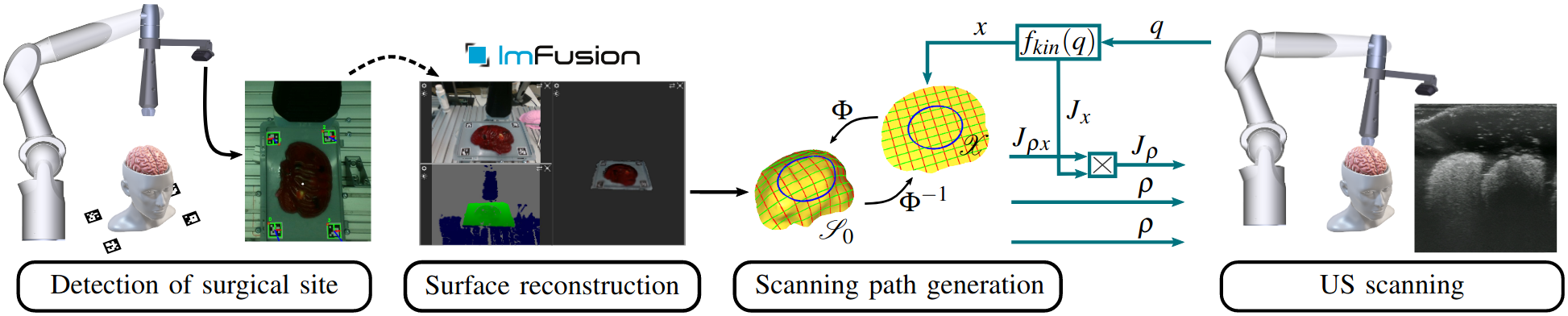}
  \caption{Workflow for our iUS tissue scanning platform. Automatic detection of the object's location and RGB-D reconstruction of its surface with a stereo camera results in a triangular mesh of the phantom. The impedance controller incorporates the geometry of this mesh into the real-time control loop, guiding the US probe to establish contact and execute scanning trajectories. Graphic based on \cite{dyck2022collab}.}
  \label{fig:workflow}
\end{figure*}

The workflow of our robotic iUS tissue scanning platform is depicted in Fig.~\ref{fig:workflow}. %The following sections introduce the individual components and their interplay.

\begin{comment}
\begin{figure*}[!ht]
  \centering
  \fontsize{9}{9}
  \selectfont
  \includesvg[width=1.\textwidth]{workflow2}
  \caption{Workflow for our iUS tissue scanning platform. Automatic detection of the object's location and RGB-D reconstruction of its surface with a stereo camera results in a triangular mesh of the phantom. The impedance controller incorporates the geometry of this mesh into the real-time control loop, guiding the US probe to establish contact and execute scanning trajectories. Graphic based on \cite{dyck2022collab}.}
  \label{fig:workflow}
\end{figure*}
\end{comment}

\subsection{Soft-Tissue-Mimicking Brain Phantoms}
A custom-made soft-tissue mimicking phantoms was made using GELITA\textsuperscript{\textregistered} GELATINE \textit{Type Ballistic 3} with a brain-like silicone mold. To mimic soft tissue and extend the durability of the phantom, the phantom contained water:glycerine:gelatine at 45:45:10~wt.\%. Inspired by \cite{Morehouse2007AdditionOM}, different objects were placed inside the phantom, including: olives, grapes, blueberries, screws. This ensures the presence of random features within the US recordings of the phantom.
%
\begin{comment}
\begin{figure}[!ht]
  \centering
  \includesvg[width=.95\columnwidth]{gelatine_phantom}
  \caption{Custom-made brain phantom with good echogenic characteristics.}
  \label{fig:phantom}
\end{figure}
\end{comment}
%

\subsection{Localisation and Surface Reconstruction}
\begin{comment}
\begin{figure}
  \centering
  \fontsize{9}{9}
  \selectfont
  \includegraphics[width=0.5\linewidth]{3d_trajectory.png}
  \caption{The setup for the localisation and reconstruction of the medical phantom. 4 ArUco markers are placed around the object. A 3D trajectory around the phantom (red dotted line) can be defined using the marker's poses.  The shown mesh is an output of the SLAM process.}
  \label{fig:sysFlow}
\end{figure}
\end{comment}
For localisation, four ArUco markers are first manually placed around the area. Then the camera is positioned above the area of interest, such that all markers are visible within the image. This is enough to initialise the automatic localisation and reconstruction routine. Using the OpenCV \cite{itseez2015opencv} ArUco library, the four fiducial markers can be identified. Using these points, a plane - defined by a centre point and normal vector - containing the phantom is calculated. Using the detected plane, the robot automatically aligns the camera such that it's z-axis points towards the plane's centre at an angle of 45\textdegree and a distance of 30~cm. Using the RGB-D reconstruction algorithm - from the \textit{ImFusion Suite} software solution by ImFusion GmbH - and executing a simple rotational trajectory - rotating the camera around the normal of the identified plane - a triangle mesh of the tissue phantom is extracted. 

\subsection{Impedance-Controlled US Scanning}

We use the impedance controller introduced in \cite{dyck2022}. 
Surface-specific coordinates are defined on the surface mesh. 
These coordinates are the distance $d\in\mathbb{R}$ between transducer and surface, and the three orientation coordinates $\boldsymbol{\epsilon}\in\mathbb{R}^{3\times1}$ that align the probe axis and surface normal. 
Two additional coordinates $(s_1, s_2)\in\mathcal{X}\subset\mathbb{R}^2$ control the position of the transducer on the surface.
We stack all coordinates in a vector $\boldsymbol{\rho}=(s_1, s_2, d, \epsilon_1, \epsilon_2, \epsilon_3)^T\in\mathbb{R}^{6\times1}$ (cmp. Fig.~\ref{fig:workflow}) and implement the unified impedance control framework by \cite{schaeffer2007}\begin{equation}\label{eq:imp}
	\boldsymbol{\tau} = \boldsymbol{J}_{\rho}^T(\boldsymbol{q})[\boldsymbol{K}_{\rho}(\boldsymbol{\rho}_d - \boldsymbol{\rho}(\boldsymbol{q}))+\boldsymbol{D}_{\rho}(\boldsymbol{q})(\boldsymbol{\dot{\rho}}_d - \boldsymbol{\dot{\rho}})].
\end{equation}
Here, $\boldsymbol{K}_{\rho},\boldsymbol{D}_{\rho}$ represent the symmetric, positive-definite stiffness and damping matrices, respectively. 
$\boldsymbol{q} \in \mathbb{R}^{7\times1}$ are the generalized configuration coordinates of the robot, ($\boldsymbol{\rho}_d, \boldsymbol{\dot{\rho}}_d$) are the desired position and velocity in the coordinates, $\boldsymbol{J}_{\rho}\in \mathbb{R}^{6\times6}$ represents the Jacobian matrix mapping joint velocities $\boldsymbol{\dot{q}}$ to velocities $\boldsymbol{\dot{\rho}}$.
Controlling the distance results in passive interaction forces between US transducer and tissue, that depend on controller and object stiffnesses.

\begin{comment}
\subsection{Scan-line entropy}

\begin{equation}
    \begin{aligned}
CE_{sl}= -\frac{1}{depth}\sum_{i=1}^{depth}[\hat{y_{sl,i}}\log(y_{sl,i} + \epsilon) \\
    + (1 - \hat{y_{sl,i}})\log(1 - y_{sl,i} + \epsilon)], \epsilon=1e-9
    \end{aligned}
\label{eq:entropy}
\end{equation}

To assure that the probe has good contact with the surface, we will take the entropy of each scan-line using Eq.~\ref{eq:entropy} - where $\hat{y_{sl}} \subset I, y_{sl}\in \Re^{depth}$ and $y_{sl}= [1, 1, ..., 1], \in \Re^{depth}$. Within the calculated "central region", an average $CE<0.5$ is required to determine. 
\end{comment}

\section{Results}
\begin{figure}[!ht]
  \centering
  \fontsize{9}{9}
  \selectfont
  \includegraphics[width=.95\columnwidth]{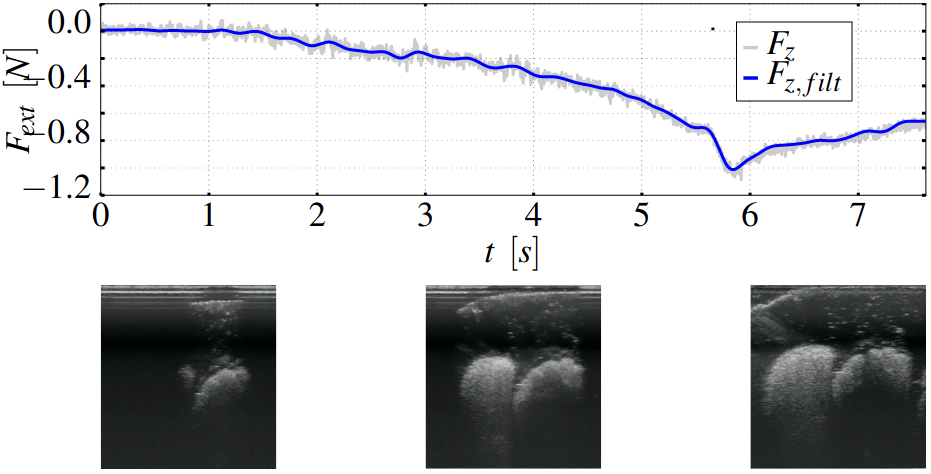}
  \caption{Interaction force between US transducer and brain phantom during contact establishment. The US images below were recorded during the interaction and show qualitatively different stages of US contact.}
  \label{fig:contact_establishment}
\end{figure}

The experimental setup can be seen in Fig.~\ref{fig:setup}. A seven degrees of freedom (DoF) DLR \textsc{Miro} surgical robotic arm is used. Attached is an Intel\textsuperscript{\textregistered} RealSense\texttrademark\ \textit{Depth Camera D435i}, and a \textit{GE 12L-RS} linear US probe. The US transducer is connected to a \textit{GE LOGIQ e} US machine. An \textit{ATI Mini45} F/T-sensor measures six DoF interaction forces during scanning. Results of the marker detection and surface 3D reconstruction can be seen in Fig.~\ref{fig:workflow}.
Fig.~\ref{fig:contact_establishment} shows the interaction force %measured during robot-guided contact establishment 
between the US probe and phantom. The US probe was aligned normal to the surface ($\boldsymbol{\varepsilon}_d=\boldsymbol{0}_{3\times1}$). The setpoint for the distance coordinate ($d_d$) was continuously decreased (increasing penetration depth) and held constant at a manually chosen value based on qualitative US image quality. %As shown in Fig.~\ref{fig:contact_establishment}, the force along the probe axis is continuously increasing as expected. 
The example US images were recorded during the experiment and show different stages of contact establishment - ranging from very little contact on the left to good probe-tissue coupling on the right. %It is important to note, that these results display merely qualitative results, as interaction force and image quality highly depend on various factors, such as controller and object stiffnesses, and echogenic characteristics of the phantom. Nonetheless, this experiment demonstrates the capabilities of our setup. 
By controlling the distance, rather than the force, safe coupling of probe and surface can be achieved - invariant to tissue properties and geometry.%, very relevant to iUS tissue scanning in brain surgery.
\begin{comment}
\begin{figure}[!ht]
  \centering
  \fontsize{9}{9}
  \selectfont
  \includesvg[width=.95\columnwidth]{force_contact}
  \caption{Interaction force between US transducer and brain phantom during contact establishment. The US images below were recorded during the interaction and show qualitatively different stages of US contact.}
  \label{fig:contact_establishment}
\end{figure}
\end{comment}

\section{Conclusion}

This work presents a framework for automated robotic iUS brain scanning. Where testing on a complex custom-made brain phantom shows that site localisation and surface reconstruction can be achieved and enables probe servoing with low interaction force. Future work will concern visual servoing for probe pose and US quality optimisation, while minimising required interaction forces.

%%%%%%%%%%%%%%%%%%%%%%%%%%%%%%%%%%%%%%%%%%%%%%%%%%%%%%%%%%%%%%%%%%%%%%%%%%%%%%%%
\bibliographystyle{IEEEtran}  
\bibliography{IEEEfull}

\end{document}